\documentclass[11pt]{article}

\usepackage[preprint]{acl}

\usepackage{times}
\usepackage{latexsym}

\usepackage[T1]{fontenc}
\usepackage{amsfonts}

\usepackage{booktabs}
\usepackage{colortbl}
\usepackage{multirow}
\usepackage{makecell}
\usepackage{array}
\usepackage{subcaption}
\usepackage{float}
\usepackage{enumitem}
\usepackage{xcolor}
\usepackage{colortbl}
\usepackage{tabularx}

\usepackage{amsmath}
\allowdisplaybreaks[4]

\usepackage[utf8]{inputenc}

\usepackage{microtype}

\usepackage{inconsolata}

\usepackage{graphicx}
\usepackage{makecell}
\usepackage{booktabs}

\usepackage{setspace}
\AtBeginDocument{
  \setlength{\abovedisplayskip}{3pt}
  \setlength{\belowdisplayskip}{3pt}
  \setlength{\abovedisplayshortskip}{2pt}
  \setlength{\belowdisplayshortskip}{2pt}
}

\setlist[itemize]{itemsep=1pt, parsep=1pt, topsep=1pt, partopsep=1pt, leftmargin=*}
\setlist[enumerate]{itemsep=1pt, parsep=1pt, topsep=1pt, partopsep=1pt, leftmargin=*}

\title{\textsc{ProgRouter}: Online Progress-Guided Orchestration for Multi-Agent LLM Workflows under Quality-Cost Tradeoffs}

\author{
Songyuan Li\textsuperscript{\ensuremath{\clubsuit},\ensuremath{\diamondsuit},}\thanks{Corresponding author. Work done while the author was with Queen Mary University of London, UK.},
Ahmed M. Abdelmoniem\textsuperscript{\ensuremath{\diamondsuit}},
Shiqiang Wang\textsuperscript{\ensuremath{\heartsuit}}
\\
\textsuperscript{\ensuremath{\clubsuit}}Aston University, UK,\,
\textsuperscript{\ensuremath{\diamondsuit}}Queen Mary University of London, UK\\
\textsuperscript{\ensuremath{\heartsuit}}University of Exeter, UK
\\
\texttt{lisy@ieee.org, ahmed.sayed@qmul.ac.uk, s.wang9@exeter.ac.uk}
}

\usepackage[most]{tcolorbox}
\usepackage{enumitem}
\begin{document}
\maketitle
\begin{abstract}
Multi-agent large language model (LLM) workflows have emerged as a powerful paradigm for solving complex, open-ended tasks through collaborative reasoning among specialized LLM agents, but they incur substantial operating costs due to repeated LLM invocations and long-horizon context accumulation. Existing cascade routing methods make one-shot, query-level decisions and cannot adapt to the dynamic, state-dependent nature of multi-step workflows, in which the right LLM at each step depends on evolving task progress, remaining task difficulty, and cost-efficiency requirements. We present \textsc{ProgRouter}, an online progress-guided routing framework that adaptively selects LLM agents across workflow steps to preserve task-solving quality while adhering to time and cost budgets. \textsc{ProgRouter} introduces a multi-view task progress scorer that combines coarse workflow outcome regimes with fine-grained signals on subtask completion, progress trends, and workflow state quality. Then, a dual-path task progress predictor and an adaptive meta-gating mechanism estimate the progress gain for each candidate routed LLM. \textsc{ProgRouter} makes online step-wise routing decisions that balance progress gain, task time budgets, and long-term operating cost efficiency. Experiments on HumanEval Plus, MBPP, MATH-500, and ASQA, spanning agentic code generation, mathematical reasoning, and retrieval-augmented long-form question answering, demonstrate that \textsc{ProgRouter} reduces the operating cost relative to key baselines while maintaining strong task-solving performance.
\end{abstract}

\section{Introduction}
Large language models (LLMs), such as ChatGPT \cite{Singh2026GPT5card} and DeepSeek \cite{Liu2025DeepSeek}, have rapidly improved their in-context reasoning and tool use, enabling LLM-based agents to support open-ended, multi-step task planning and problem solving \cite{Deng2025Agentpro}. Multi-agent LLM workflows \cite{Zhang2025Swarm, Yao2023React, Fourney2024Magentic} further extend this capability by coordinating specialized agents to decompose complex queries and solve subtasks through iterative reasoning. However, these workflows are costly to operate, as they require repeated LLM calls across multiple steps and maintain long-horizon task contexts, leading to high token consumption, compute usage, energy demand, and latency \cite{Lin2026Stop, Xiao2026agentdiet}. Therefore, developing sustainable, cost-aware serving strategies, especially by selecting appropriate LLMs for different agents and workflow steps, is crucial for improving cost efficiency while preserving task-solving performance.

Cascade LLM serving \cite{Jiang2026Cascadia, Ong2025Route, Chen2024RouterDC, Woisetschlaeger2025MESS} offers a promising direction by routing user queries to LLMs of varying sizes and capabilities based on task complexity. Existing methods assign simpler queries to lightweight LLMs while reserving stronger ones for harder requests, often with minimal quality degradation. However, these \textit{one-shot} routing methods do not adequately support multi-agent workflows, which require not a single upfront decision but a sequence of coordinated choices about which agents to call, when to call them, and how to combine their complementary strengths, adapting online to the evolving workflow state. This gap motivates the fundamental question:
\vspace{-1.6em}
\begin{tcolorbox}[
    colback=gray!7,
    colframe=black!40,
    boxrule=0.5pt,
    arc=2mm,
    left=1.5mm,
    right=1.5mm,
    top=1.5mm,
    bottom=1.5mm
]%
\textit{%
How to dynamically route LLM agents across workflow steps to preserve task-solving quality under operating cost budgets?
}
\end{tcolorbox}
\vspace{-0.75em}
\noindent Answering this question is challenging because:
\begin{itemize}[leftmargin=*]
\item Multi-agent LLM workflows are open-ended and stateful. The ``right'' agent at each step depends on the current task progress, partial solution quality, and emergent collaboration needs. Traditional LLM selection methods, which rely on static benchmarks \cite{Naveed2025Comprehensive} or offline human-preference datasets \cite{Ong2025Route}, cannot meet the online, in-context routing demands of multi-step agentic workflows.
\item Effective LLM agent routing requires accurately assessing how much task progress has been made and how much difficulty remains. Yet intermediate workflow states are often noisy and partially resolved, with key progress factors hidden or implicit \cite{Ma2024Agentboard, Li2026Spinning}. Miscalibrated estimates either waste operating cost budget on unnecessary strong-agent calls or overuse weak agents, causing irrecoverable quality degradation \cite{Deng2025Agentpro}.
\item LLM agent routing must respect the workflow's remaining cost budget, but future steps and subtask difficulties are unknown a priori. Greedy policies that always select the strongest LLM may exhaust the budget early, while conservative policies under-utilise strong agents. This calls for principled online algorithms that balance task progress against long-term operating cost.
\end{itemize}

\noindent\textbf{Contributions.} To address these challenges, this paper presents \textsc{ProgRouter}, an online progress-guided routing framework that adaptively selects LLM agents across workflow steps under time and cost budgets, while preserving strong task-solving quality. \textsc{ProgRouter} continuously learns workflow progress patterns from diverse routing trajectories as multi-agent LLM workflows execute, and progressively refines its routing strategy. At its core, \textsc{ProgRouter} introduces a multi-view task progress scorer that combines coarse workflow outcome regimes with fine-grained signals on subtask completion, task progress trends, and workflow state quality. This task progress scorer serves as a lightweight domain adapter that converts observable workflow signals into a unified progress representation, while the subsequent dual-path progress predictor design principles and online quality-cost-aware routing algorithm remain transferable across different agentic domains. Building on this, a dual-path progress predictor with structured and semantic paths and an adaptive meta-gating mechanism estimates the progress gain of each candidate routed LLM. \textsc{ProgRouter} makes online step-wise LLM agent routing decisions that balance task progress gain, task time budgets, and long-term operating cost efficiency, enabling workflow operation that is quality-driven, deadline-aware, and cost-efficient.

\begin{figure*}[htbp]
  \centering
  \vspace{-1ex}
  \includegraphics[width=0.805\linewidth]{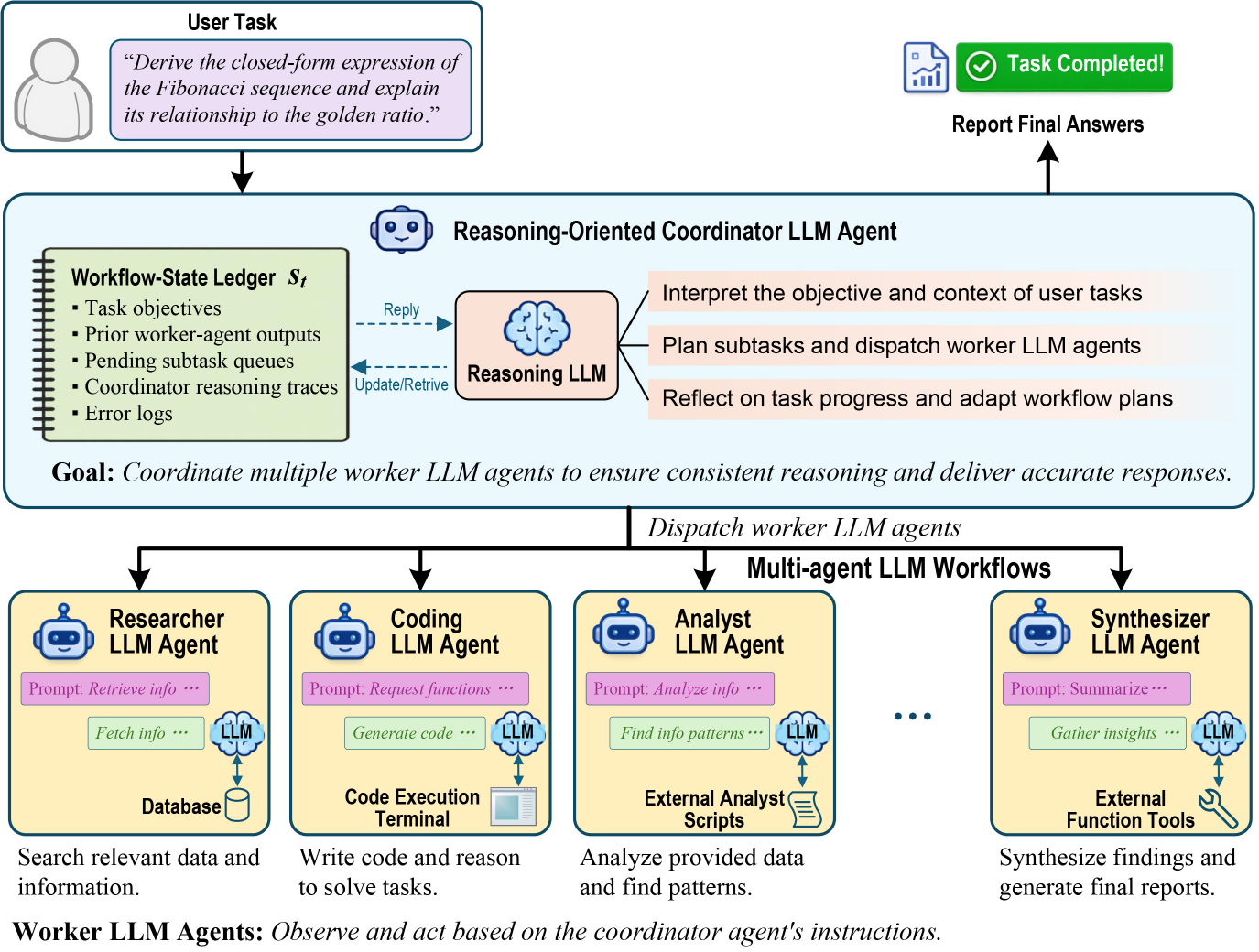}
  \vspace{-0.5em}
  \caption{Overview of LLM agent orchestration in collaborative multi-agent LLM workflows.}
  \label{fig:system-model}
  \vspace{-1.25em}
\end{figure*}

\section{Problem Setting}\label{sec:prob_setting}
\subsection{Collaborative Multi-agent LLM Workflow}
As shown in Figure \ref{fig:system-model}, a multi-agent LLM workflow $w=\{C, \mathcal{R}, \mathcal{M}\}$ solves a user task by executing a sequence of $T$ agent dispatch steps:
\begin{itemize}[leftmargin=*]
\item $\mathcal{R}$ denotes the set of predefined worker LLM agent roles $r$ (e.g., analyst and coder).
\item $C$ denotes a coordinator LLM agent responsible for subtask planning, worker LLM agent dispatch, and workflow adaptation during execution.
\item $\mathcal{M}$ denotes the set of available LLMs $m$ of different sizes and capacities from which models are to instantiate an agent with role $r\in\mathcal{R}$. Let $\mathcal{M}_r\subset\mathcal{M}$ be the subset of LLMs eligible to serve agent role $r$.
\end{itemize}
\noindent During operation, the multi-agent LLM workflow $w$ maintains the evolving \textit{workflow state} $s$ indicating the problem-solving status. Each agent dispatch step $t\in\{1\cdots T\}$ invokes the agent role $r_t$ and yields an updated workflow state $s_{t+1}$, contributing to the final normalized task-performance score $\mathcal{P}(w)\in[0,1]$.

\subsection{LLM Agent Orchestration}
Different LLMs have distinct strengths and limitations \cite{Fourney2024Magentic, Zhang2025Swarm}, and the goal of LLM agent routing is to leverage their complementary capabilities. Such heterogeneous workflows include structured tasks such as code generation and mathematical reasoning, as well as open-ended workflows such as retrieval-augmented question answering, where task progress must be inferred from intermediate evidence synthesis, retrieval results, and answer refinement. For example, an LLM with programming capabilities is selected for code-generation subtasks, while a reasoning-oriented LLM could be assigned to higher-level analytical tasks, e.g., dynamic subtask orchestration and critic-based evaluation of intermediate agent outputs. Therefore, the effective orchestration of LLM agents within a workflow $w$ depends on the domain and difficulty of the dispatched tasks. To achieve this, we employ \textit{a ledger-driven coordinator agent} $C$ for dynamic task planning and worker-agent role dispatch, and an \textit{online progress-guided LLM routing algorithm} for adaptive agentic LLM selection.

\textbf{Ledger-Driven Coordinator Agent.}
The coordinator agent $C$, powered by a reasoning-oriented LLM, decomposes high-level user objectives into executable subtasks and sequentially allocates them to specialised worker roles $r\in\mathcal{R}$. Unlike static rule-based planners, $C$ maintains structured workflow-state ledgers $s_t$ that summarise task objectives, current task progress, intermediate reasoning traces, and prior worker-agent outputs. These ledgers enable continuous monitoring of workflow states and detection of execution failures. Leveraging information recorded in the ledgers, the coordinator LLM reasons over workflow states and dynamically revises execution plans through subtask reallocation or coordination adjustments. This facilitates adaptive planning and error recovery in complex, long-horizon workflows.

\textbf{Progress-Guided LLM Routing Algorithm.} 
While the coordinator agent $C$ determines \emph{which worker agent role} should be dispatched next, we expect a progress-guided LLM routing algorithm that determines \emph{which LLM instance} should instantiate the selected role. Specifically, given the real-time workflow execution state maintained in coordinator ledgers, the online routing algorithm evaluates subtask characteristics, including the dispatched subtask domain, estimated subtask difficulty, and historical execution performance, to select suitable LLMs from the candidate set $\mathcal{M}_r \subseteq \mathcal{M}$ for the worker agent role~$r$. This adaptive routing strategy aims to balance LLM capability requirements against time and operating cost throughout workflow execution. We implement it as \textsc{ProgRouter} whose detailed methodology is presented below.

\begin{figure*}[htbp]
  \centering
  \vspace{-1ex}
  \includegraphics[width=0.776\linewidth]{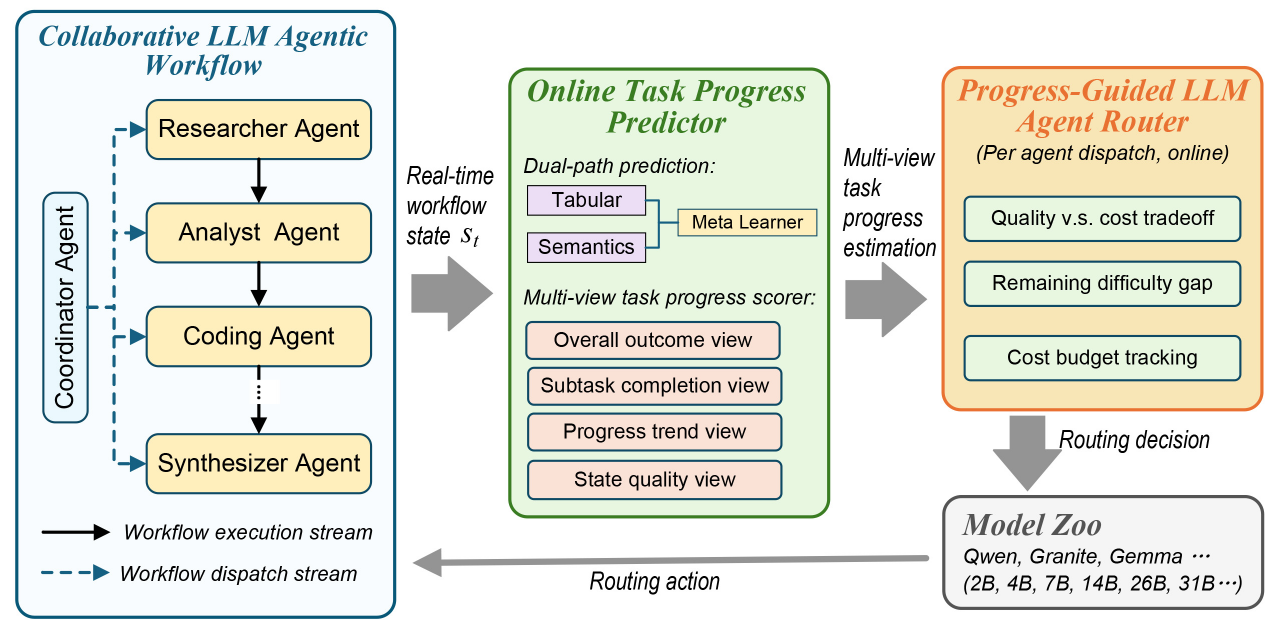}
  \vspace{-0.5em}
  \caption{Overall procedure of \textsc{ProgRouter}. A coordinator LLM agent manages a collaborative multi-agent LLM workflow by dispatching worker agents and observing real-time workflow states. The online task progress predictor estimates \textit{multi-view task progress} from \textit{structured} and \textit{semantic} state representations, and the progress-guided router selects suitable LLMs from the model zoo by jointly considering predicted progress gain, remaining task difficulty, and cost-budget constraints.}
  \label{fig:framework}
  \vspace{-1.25em}
\end{figure*}

\section{ProgRouter Methodology}\label{sec:method}

\subsection{LLM Routing Control Problem}\label{subsec:llm_routing_control}
Figure~\ref{fig:framework} illustrates the overall procedure of \textsc{ProgRouter}. We consider a multi-agent LLM workflow that continuously serves a stream of $W$ user tasks. For each user task $w\in\{1,\cdot,W\}$, 
the workflow executes a sequence of $T_w$ agent dispatch steps.  Different user tasks exhibit varying complexity and service requirements, resulting in the constraints of task time $\Gamma_w$ and operating costs $E_w$. Besides, we impose a long-term averaging operating cost constraint $\widetilde{E}$ to regulate overall system-level cost efficiency across multiple served tasks.

We seek an online adaptive LLM agent orchestration strategy $\pi(s_t,r_t)\mapsto m_t$ across agent dispatch steps. Based on the current workflow state $s_t$, each step $t$ selects a suitable LLM $m_t\in\mathcal{M}_{r_t}$ for the dispatched agent role $r_t$:
\begin{subequations}
\begin{align}
& \max\limits_{\pi} \; \textstyle\frac{1}{W} \sum\limits_{w=1}^W \mathcal{P}\left(w\right) \label{eq:obj} \\
&\textit{ s.t. }\; \textstyle\sum\limits_{t=1}^{T_w} E(m_t) \leq E_w, \forall\, w \in\{1\cdots W\}, \label{eq:energy_constraint}\\
&\quad\;\;\;\,\textstyle\sum\limits_{t=1}^{T_w} \Gamma(m_t) \leq \Gamma_w, \forall\, w \in\{1\cdots W\}, \label{eq:time_constraint}\\
&\quad\;\;\;\textstyle\frac{1}{W} \sum\limits_{w=1}^W\, \sum\limits_{t=1}^{T_w}\; E(m_t) \leq \widetilde{E}, \label{eq:soft_constraint}
\end{align}
\end{subequations}
where $E(m_t)$ and $\Gamma(m_t)$ denote the operating cost and execution time incurred by the selected LLM $m_t$, respectively. Formally, $\pi(s_t,r_t)$ aims to maximize the workflow's task-solving performance while satisfying both task-specific and long-term averaging operating cost constraints. This frames LLM agent orchestration as an online constrained optimization process over long-horizon, multi-task workflow execution.

\textbf{Challenges.} Our online control problem is challenging for two main reasons. \textit{First,} the task-solving performance $\mathcal{P}(w)$ can be usually observed after the entire workflow completes. This makes it non-trivial to assess the contribution to progress of each agent dispatch step's LLM routing decisions in real time. \textit{Second,} optimizing the system-level operating costs involves a time average, which is difficult to estimate a priori because future user tasks and agent dispatch steps are heterogeneous and unknown in advance.

\subsection{Multi-View Task Progress Representation}
\label{sec:progress_encoding}
To enable online, progress-guided LLM routing, \textsc{ProgRouter} encodes intermediate workflow states into a multi-view task-progress representation. Given the current workflow state $s_t$, the multi-view task progress scorer computes a normalized progress score $g(s_t) \in [0,1]$. It quantifies how close the current workflow is to successful task completion. Higher values of $g(s_t)$ indicate greater progress. To achieve robust and fine-grained progress estimation across heterogeneous workflow states, we evaluate the workflow from multiple complementary perspectives:
\vspace{-0.25em}\begin{itemize}[leftmargin=*]
\item \textbf{Overall outcome view}: Assesses the overall health and validity of the current workflow state. It maps $s_t$ into one of four coarse regimes $c_t\in\{\textup{invalid, recoverable, partial success, complete}\}$ using rule-based conditions over observable workflow-state signals. This produces the base score $b(c_t)$. 

\item \textbf{Subtask completion view}: Measures the degree to which the user's subtasks, constraints, and other evaluation criteria have been satisfied. It computes the fraction of currently fulfilled task progress requirements among all requirements maintained in the workflow state $s_t$, based on coordinator-ledger states and worker-agent execution feedback, yielding the sub-score $r_t$.

\item \textbf{Progress trend view}: Captures short-term progress dynamics by analyzing whether the workflow is improving, stagnating, or regressing over recent steps. It is computed algorithmically from recent changes in observable progress signals maintained in the workflow state $s_t$, specifically as the average progress delta over the last few steps, producing the sub-score $d_t$.

\item \textbf{State quality view}: Evaluates the semantic and structural meaningfulness of the latest workflow-state transition. It combines two lightweight signals: (1) semantic similarity between consecutive workflow-state summaries, computed using a lightweight sentence-embedding encoder (e.g., MiniLM~\cite{Wang2021MiniLMv2}); and (2) structural workflow-state differences (e.g., newly resolved subtasks, newly retrieved evidence, or newly detected failures). These signals are computed directly from existing workflow states, yielding the sub-score $e_t$.

\end{itemize}

\noindent Finally, the multi-view task progress score $g(s_t)$ is obtained through \textit{hierarchical aggregation}:
\begin{equation}
    g(s_t) = b(c_t) + \alpha_{c_t} r_t + \beta_{c_t} d_t + \gamma_{c_t} e_t,
\end{equation}
where $b(c_t)$ is the base score associated with the coarse outcome regime, and $\alpha_{c_t}, \beta_{c_t}, \gamma_{c_t} \geq 0$ are weighting coefficients. Note that this progress scorer provides a lightweight domain adapter by algorithmically mapping domain-specific observable workflow milestones into a common multi-view task-progress representation. For a new agentic task domain, only the observable milestone definitions and coarse outcome regimes need adaptation. The downstream online exploration-and-update procedure and online budget-aware LLM agent routing algorithm remain domain-independent across task domains.

\textbf{Key Insight.} The hierarchical multi-view scorer design combines a reliable coarse anchor (overall outcome) with fine-grained signals from three complementary analytical views, enabling the detection of both obvious and subtle task progress. It generates dense step-wise supervisory signals that are more timely and informative than a sparse final workflow outcome for guiding LLM agent routing during workflow execution.

\subsection{Online Task Progress Predictor}
\textsc{ProgRouter} further requires \textit{a forward-looking progress signal} to determine which agentic LLM should be invoked next to effectively advance task completion. The gain in progress of an LLM routing decision depends not only on the LLM's size and capacity but, more critically, on the current workflow state and the remaining task difficulty. A powerful LLM yields substantial improvement when the workflow is blocked by unresolved errors, yet provides limited marginal benefit once the user task is near completion. Therefore, accurate prediction of task progress requires a precise understanding of the current workflow state.

\textbf{Dual-Path Task Progress Prediction.} Given the current workflow state $s_t$ and a candidate LLM $m_t\in\mathcal{M}_{r_t}$, the task progress predictor estimates the step-wise progress gain when invoking $m_t$ at the agent dispatch step $t$:
\begin{equation}
    \hat{y}_{t}=P_{\Theta}(s_t,m_t)\textup{,}
\end{equation}
which facilitates intermediate progress-guided online LLM routing. Here, $s_t$ is represented through complementary structured and semantic features. To obtain reliable progress estimates across heterogeneous workflow regimes $s_t$, we implement $P_{\Theta}$ as a \emph{dual-path predictor} with two complementary prediction paths and an adaptive meta-gating mechanism:
\begin{itemize}[leftmargin=*]
\item \textbf{Structured path:} 
Operates on a tabular feature vector $x_{t}^{\mathrm{str}} = \phi_{\mathrm{str}}(s_t, m_t)$ that encodes explicit workflow signals, including the current multi-view task progress score $g(s_t)$, its constituent signals (e.g., subtask completion and recent progress trends), LLM agent dispatch history, and the candidate LLM routing decision $m_t$. It produces a progress-gain estimate:
\begin{equation}
\hat{y}_{t}^{\mathrm{str}}
=
P_{\mathrm{str}}\!\big(x_{t}^{\mathrm{str}}\big),
\end{equation}
where $P_{\mathrm{str}}$ represents a tree-based regressor (e.g., random forest, XGBoost) that excels at learning heterogeneous, low-dimensional tabular features.

\item \textbf{Semantic path:} 
Operates on a compact natural-language summary of the workflow state $s_t$ and the candidate routing decision $m_t$, produced by the coordinator LLM agent $C$. The summary is encoded into a dense language embedding vector
$x_{t}^{\mathrm{sem}} = \phi_{\mathrm{sem}}(s_t, m_t)$
by a lightweight sentence-embedding model $\phi_{\mathrm{sem}}$ (e.g., MiniLM \cite{Wang2021MiniLMv2}),
yielding a progress-gain estimate:
\begin{equation}
\hat{y}_{t}^{\mathrm{sem}}
=
P_{\mathrm{sem}}\!\big(x_{t}^{\mathrm{sem}}\big),
\end{equation}
where $P_{\mathrm{sem}}$ is likewise a tree-based
regressor. The semantic path captures subtle, qualitative
workflow cues that are difficult to express through tabular
features.
\end{itemize}

The dual-path estimates $\hat{y}_{t}^{\mathrm{str}}$ and $\hat{y}_{t}^{\mathrm{sem}}$ are finally combined through a tree-based \emph{meta-gated} learner (e.g., XGBoost) that adaptively routes trust between the two prediction paths based on the online workflow state $s_t$:
\begin{equation}
\hat{y}_{t}
=
P_{\mathrm{meta}}
(
\hat{y}_{t}^{\mathrm{str}},\;
\hat{y}_{t}^{\mathrm{sem}}
)\textup{.}
\end{equation}
\noindent At each agent dispatch step $t$, the dual-path predictor evaluates online all candidate LLMs $m_t \in \mathcal{M}_{r_t}$ with negligible overhead, as both prediction paths rely on a lightweight sentence encoder or moderate tree-based regressors. The resulting progress-gain estimate $\hat{y}_t(m_t)$ is incorporated into the routing objective of the coordinator LLM agent $C$, enabling worker-agent dispatch decisions that jointly optimize predicted task-progress gain against time and operating cost.

\textbf{Key Insight.} Owing to its meta-gated design, our dual-path task progress predictor adaptively combines structured and semantic evidence under different workflow regimes. When explicit workflow progress signals are reliable, it can place more emphasis on the structured estimate. On the other hand, when the workflow state contains subtle progress bottlenecks and sparse observable feedback, it can rely more on the semantic estimate.

\subsection{Online Decision Making}
Building on the dual-path task progress predictor, \textsc{ProgRouter} introduces an online decision-making algorithm that turns the predicted progress gain $\hat{y}_t(m_t)$ into  LLM routing actions at each agent dispatch step $t$.

\textbf{Virtual Cost Queues.} Our approach is inspired by the Lyapunov drift-plus-penalty framework \cite{Neely2010Stochastic}. We capture the system operating cost-efficiency violation, i.e., accumulated overshooting of $\widetilde{E}$, in a \textit{virtual queue} $Q$ with the following update procedure after completing the user task $w$:
\begin{equation}
\textstyle Q_{w+1}=\max\{0, Q_{w} + \sum\nolimits_{t=1}^{T_w}\; E(m_t) - \widetilde{E}\}
\end{equation}
where we initialize $Q_1=0$. Intuitively, this captures a violation of the long-term averaging operating-cost constraint (\ref{eq:soft_constraint}). Thus, we aim to maximize the workflow performance of every user task $w$ while minimizing the virtual queue length. The quality-cost trade-off is formulated as below, where  $V$ is the trade-off coefficient:
\begin{subequations}
\begin{align}
& \textstyle\max\limits_{\pi} \; V\!\cdot\!\mathcal{P}\left(w\right) - Q_w\left(\sum\limits_{t=1}^{T_w}\; E(m_t) - \widetilde{E}\right) \\
& \;\textit{ s.t. }\; \textup{Constraints (\ref{eq:energy_constraint}) and (\ref{eq:time_constraint}).}
\end{align}
\end{subequations}

\textbf{Online Step-Wise Routing Decision.}
At each agent step $t$ in the workflow $w$, \textsc{ProgRouter} selects the LLM $m_t\!\in\!\mathcal{M}_{r_t}$ that best advances the task under the time and operating cost budget. The routing decision is made immediately and irrevocably without knowledge of future agent dispatch steps or workflow evolution.

\textit{Progress-gap-aware LLM routing.}
We value the task-solving capacity of candidate LLM $m_t$ as:
\begin{equation}
\widetilde{P}(m_t, s_t) = V \cdot (1 - g(s_t)) \cdot \hat{y}_t(m_t)
\end{equation}
The factor $(1-g(s_t))$ captures the intuition that the task progress gain $\hat{y}_t(m_t)$ is more valuable when the user task is still far from completion. It would incentivize \textsc{ProgRouter} to select stronger LLM agents when the workflow has substantial room for progress, where additional progress is expected to yield the larger benefit.
  
\textit{Budget-aware cost penalty.} In addition to the system-level virtual cost queue, \textsc{ProgRouter} respects the task-specific time and operating cost budgets ($\Gamma_w$ and $E_w$). When a workflow has already consumed a large portion of its cost budget, the router should be more conservative in selecting costly LLMs. We define the budget-aware cost coefficients as:
\begin{equation}
\textstyle c_t^{\Gamma} = \exp\!\left( \frac{\Gamma(m_t)+\sum_{i<t}\Gamma(m_i)}{\Gamma_w}\right)
\end{equation}
\begin{equation}
\textstyle c_t^{E} = \exp\!\left(\frac{E(m_t)+\sum_{i<t}E(m_i)}{E_w} \right)
\end{equation}
These coefficients increase as the cumulative time and operating cost consumption approach their corresponding budgets. Therefore, the same invoked LLM agent receives a larger cost penalty at later stages if the workflow has already consumed more budgets. This encourages \textsc{ProgRouter} to preserve budget headroom and avoid repeatedly selecting costly LLMs unless their predicted progress gain is sufficiently large. The final online LLM routing score is:
\begin{equation}
\begin{aligned}
    \mathrm{score}(m_t)
    = & \widetilde{P}(m_t)
    - Q_w \cdot (E(m_t)-\widetilde{E}) \\
    &
    - c_t^\Gamma \cdot \Gamma(m_t) - c_t^E\cdot E(m_t).
\end{aligned}
\label{eq:score}
\end{equation}
At each agent dispatch step, \textsc{ProgRouter} selects
$m_t^\star = \arg\max_{m_t\in\mathcal{M}_{r_t}}\mathrm{score}_t(m_t)$.
The routing score balances three factors: task progress, system-level operating cost-efficiency pressure, and per-task time/operating cost budget consumption. This allows \textsc{ProgRouter} to select stronger LLMs when they are expected to bring meaningful progress, while avoiding excessive time and operating costs during online workflow execution.   

\vspace{-0.2em}\textbf{Online Predictor Learning.}
Our task progress predictor $P_{\Theta}$ is learned online through an exploration-and-update procedure. At each agent dispatch step, with probability $\varepsilon$, \textsc{ProgRouter} randomly selects a candidate LLM to collect an unbiased training sample; with probability $1-\varepsilon$, it selects the LLM with the highest routing score in Eq.~(\ref{eq:score}) using the current task progress predictor. After the selected agent step is executed, the realized progress gain is computed as $y_t = g(s_{t+1}) - g(s_t)$, and exploration samples are added to the training buffer $\mathcal{D}$. Therefore, the multi-view task progress score $g(\cdot)$ plays two complementary roles: $g(s_t)$ characterizes the current workflow progress as part of the predictor input, while the realized change $g(s_{t+1})-g(s_t)$ provides the online supervision signal after execution. Once enough samples have been collected, the predictor is periodically updated and used to guide future LLM routing decisions. As more online training samples accumulate, $\varepsilon$ gradually decays, shifting the router from exploration toward predictor-guided exploitation. In this way, \textsc{ProgRouter} adaptively improves LLM routing without offline training data, leading to more accurate task progress estimation and better overall workflow performance.

\begin{table*}[!t]
\centering
\caption{Main benchmark results on HumanEval Plus~\cite{Liu2023Code}, MBPP~\cite{Austin2021Program}, MATH-500~\cite{Hendrycks2021Measuring}, and ASQA datasets~\cite{Stelmakh2022ASQA}. \textcolor{green!60!black}{Green} highlights all methods that satisfy the long-term system operating cost efficiency requirements $\widetilde{E}$, and \textcolor{red!95!black}{red} values indicate violations. \textbf{Bold} denotes the best value, and \underline{underlining} indicates the second-best value for task completion rate (Pass/Precision), energy consumption, and workflow execution time among methods satisfying $\widetilde{E}$. We report the agentic workflow's operating cost in energy consumption (Joule).}
\vspace{-0.95em}
\label{tab:results_all}
\small
\setlength{\tabcolsep}{4pt}
\renewcommand{\arraystretch}{1}

\par\smallskip
\begin{tabularx}{\linewidth}{@{}X ccc | cccc | ccccc @{}}
\toprule
\textbf{(a) HumanEval Plus Dataset}
& \multicolumn{3}{c|}{\textbf{Performance}}
& \multicolumn{4}{c|}{\textbf{Qwen 2.5-Coder (\%)}}
& \multicolumn{5}{c}{\textbf{Qwen 3.5 (\%)}} \\
\cmidrule(lr){2-4}\cmidrule(lr){5-8}\cmidrule(l){9-13}
\textit{Methods} ($\widetilde{E}$ = 4,800 J)
& \makecell{Pass\\(in \%)$\uparrow$}
& \makecell{Energy\\(in J)$\downarrow$}
& \makecell{Time\\(in sec)$\downarrow$}
& 0.5B & 7B & 14B & 32B
& 2B & 4B & 9B & 27B & 35B \\
\midrule
Qwen2.5-Coder 0.5B Only & 19.1 & \textcolor{red!95!black}{5,952}  & 21.0 & 100 & --  & --  & --  & --  & --  & --  & --  & -- \\
Qwen2.5-Coder 32B Only  & 94.0 & \textcolor{red!95!black}{7,837}  & 17.5 & --  & --  & --  & 100 & --  & --  & --  & --  & -- \\
Qwen 3.5 2B Only        & 78.3 & \textcolor{red!95!black}{5,475}  & 19.2 & --  & --  & --  & --  & 100 & --  & --  & --  & -- \\
Qwen 3.5 35B Only       & 97.1 & \textcolor{red!95!black}{5,443}  & 19.7 & --  & --  & --  & --  & --  & --  & --  & --  & 100 \\
\midrule
\textit{Educated Guessing}                 & 91.5             & \textcolor{red!95!black}{4,916}               & 15.1             & 6.3  & 92.1 & --  & 1.6 & --   & --  & --  & --  & -- \\
\textit{CASCADIA}~\cite{Jiang2026Cascadia} & 84.8             & \textcolor{green!60!black}{\underline{4,658}}  & 16.6             & 81.3 & --   & --  & --  & 6.2  & 12.5 & -- & --  & -- \\
\textit{MasRouter}~\cite{Yue2025Masrouter} & \underline{90.9} & \textcolor{green!60!black}{\textbf{4,483}}    & \textbf{13.2}    & 59.4 & 25.0 & --  & 6.3 & 1.6  & 3.1 & 1.4 & 1.6 & 1.6 \\
\rowcolor{gray!25}
\textbf{\textsc{ProgRouter}} (\textit{ours}) & \textbf{93.0}  & \textcolor{green!60!black}{4,796}             & \underline{13.7} & 84.3 & 3.1  & 4.7 & 3.1 & 1.6  & --  & --  & 1.6 & 1.6 \\
\bottomrule
\end{tabularx}

\vspace{0.6em}

\begin{tabularx}{\linewidth}{@{}X ccc | cccc | ccccc @{}}
\toprule
\textbf{(b) MBPP Dataset}
& \multicolumn{3}{c|}{\textbf{Performance}}
& \multicolumn{4}{c|}{\textbf{Qwen 2.5-Coder (\%)}}
& \multicolumn{5}{c}{\textbf{Qwen 3.5 (\%)}} \\
\cmidrule(lr){2-4}\cmidrule(lr){5-8}\cmidrule(l){9-13}
\textit{Methods} ($\widetilde{E}$ = 4,500 J)
& \makecell{Pass\\(in \%)$\uparrow$}
& \makecell{Energy\\(in J)$\downarrow$}
& \makecell{Time\\(in sec)$\downarrow$}
& 0.5B & 7B & 14B & 32B
& 2B & 4B & 9B & 27B & 35B \\
\midrule
Qwen2.5-Coder 0.5B Only &  8.0 & \textcolor{red!95!black}{5,571}  & 22.3 & 100 & --  & --  & --  & --  & --  & --  & --  & -- \\
Qwen2.5-Coder 32B Only  & 85.3 & \textcolor{red!95!black}{7,207}  & 17.0 & --  & --  & --  & 100 & --  & --  & --  & --  & -- \\
Qwen 3.5 2B Only        & 64.5 & \textcolor{red!95!black}{14,308} & 47.6 & --  & --  & --  & --  & 100 & --  & --  & --  & -- \\
Qwen 3.5 35B Only       & 93.9 & \textcolor{red!95!black}{4,804}  & 16.4 & --  & --  & --  & --  & --  & --  & --  & --  & 100 \\
\midrule
\textit{Educated Guessing}                 & 65.2             & \textcolor{green!60!black}{3,831}             & \underline{11.5} & 94.8 & 1.3 & --  & 1.3 & --   & 1.3 & 1.3 & --  & -- \\
\textit{CASCADIA}~\cite{Jiang2026Cascadia} & \underline{78.5} & \textcolor{green!60!black}{3,857}             & 13.3             & 77.6 & 1.3 & --  & --  & 19.8 & 1.3 & --  & --  & -- \\
\textit{MasRouter}~\cite{Yue2025Masrouter} & 67.8             & \textcolor{green!60!black}{\underline{3,700}}             & \underline{11.5} & 1.3  & 19.2 & -- & --  & 7.7  & 1.3 & 20.5 & 1.3 & 48.7 \\ \rowcolor{gray!25}
\textbf{\textsc{ProgRouter}} (\textit{ours}) & \textbf{79.4}  & \textcolor{green!60!black}{\textbf{3,376}}   & \textbf{10.3}    & 78.5 & 8.9 & 3.7 & 1.3 & 1.3  & 3.7 & --  & 1.3 & 1.3 \\
\bottomrule
\end{tabularx}

\vspace{0.6em}

\begin{tabularx}{\linewidth}{@{}X ccc | cccc | ccccc @{}}
\toprule
\textbf{(c) MATH-500 Dataset}
& \multicolumn{3}{c|}{\textbf{Performance}}
& \multicolumn{4}{c|}{\textbf{Granite 4.1 (\%)}}
& \multicolumn{5}{c}{\textbf{Gemma 4 (\%)}} \\
\cmidrule(lr){2-4}\cmidrule(lr){5-8}\cmidrule(l){9-13}
\textit{Methods} ($\widetilde{E}$ = 7,000 J)
& \makecell{Pass\\(in \%)$\uparrow$}
& \makecell{Energy\\(in J)$\downarrow$}
& \makecell{Time\\(in sec)$\downarrow$}
& 3B & 8B & 30B & \phantom{32B}
& 2B & 4B & 26B & 31B & \phantom{35B} \\
\midrule
Granite 4.1 3B Only  & 43.4 & \textcolor{red!95!black}{8,830}  & 28.1  & 100 & --  & --  & \textcolor{white}{--} & --  & --  & --  & --  & \textcolor{white}{--} \\
Granite 4.1 30B Only & 89.0 & \textcolor{red!95!black}{11,950} & 29.8  & --  & --  & 100 & \textcolor{white}{--} & --  & --  & --  & --  & \textcolor{white}{--} \\
Gemma 4 2B Only     & 67.1 & \textcolor{red!95!black}{11,974} & 43.8  & --  & --  & --  & \textcolor{white}{--} & 100 & --  & --  & --  & \textcolor{white}{--} \\
Gemma 4 31B Only     & 92.2 & \textcolor{red!95!black}{26,276} & 55.09 & --  & --  & --  & \textcolor{white}{--} & --  & --  & --  & 100 & \textcolor{white}{--} \\
\midrule
\textit{Educated Guessing}                 & 79.3             & \textcolor{red!95!black}{7,023}               & 20.9             & --   & 98.8 & --  & \textcolor{white}{--} & --   & 1.2  & --  & --  & \textcolor{white}{--} \\
\textit{CASCADIA}~\cite{Jiang2026Cascadia} & \textbf{87.8}    & \textcolor{green!60!black}{\underline{6,875}}             & \underline{24.6} & 7.5  & 0.5  & 0.5 & \textcolor{white}{--} & 88.5 & 2.0  & 0.5 & 0.5 & \textcolor{white}{--} \\
\textit{MasRouter}~\cite{Yue2025Masrouter} & 73.6             & \textcolor{red!95!black}{8,294}               & 24.0             & 72.0 & 7.6  & --  & \textcolor{white}{--} & 6.4  & 10.2 & 1.3 & 2.5 & \textcolor{white}{--} \\
\rowcolor{gray!25}
\textbf{\textsc{ProgRouter}} (\textit{ours}) & \underline{84.3} & \textcolor{green!60!black}{\textbf{6,112}} & \textbf{19.0}    & 91.0 & 5.1  & --  & \textcolor{gray!25}{--} & --   & --   & 2.4 & 1.5 & \textcolor{gray!25}{--} \\
\bottomrule
\end{tabularx}

\vspace{0.6em}

\begin{tabularx}{\linewidth}{@{}X ccc | cccccc | ccc @{}}
\toprule
\textbf{(d) ASQA Dataset}
& \multicolumn{3}{c|}{\textbf{Performance}}
& \multicolumn{6}{c|}{\textbf{Qwen 3.5 (\%)}}
& \multicolumn{3}{c}{\textbf{Qwen 3.6 (\%)}} \\
\cmidrule(lr){2-4}
\cmidrule(lr){5-10}
\cmidrule(l){11-13}

\textit{Methods} ($\widetilde{E}=19,000\,\mathrm{J}$)
& \makecell{Precision\\(in\%)$\uparrow$}
& \makecell{Energy\\(in J)$\downarrow$}
& \makecell{Time\\(in sec)$\downarrow$}
& 2B
& 4B
& 9B
& 27B
& 35B
& \phantom{35B}
& 27B
& 35B
& \phantom{35B}
\\
\midrule

Qwen 3.5 2B Only
& 89.2 & \textcolor{green!60!black}{\underline{17,028}} & \underline{60.2}
& 100 & -- & -- & --
& --
& \phantom{--} & -- & -- & \phantom{--}
\\

Qwen 3.5 4B Only
& 88.0 & \textcolor{red!95!black}{22,400} & 70.3
& -- & 100 & -- & --
& --
& \phantom{--} & -- & -- & \phantom{--}
\\

Qwen 3.5 9B Only
& 86.5 & \textcolor{red!95!black}{28,617} & 83.5
& -- & -- & 100 & --
& --
& \phantom{--} & -- & -- & \phantom{--}
\\

Qwen 3.5 27B Only
& 90.0 & \textcolor{red!95!black}{34,000} & 92.2
& -- & -- & -- & 100
& --
& \phantom{--} & -- & -- & \phantom{--}
\\

Qwen 3.5 35B Only
& 90.5 & \textcolor{red!95!black}{36,400} & 98.1
& -- & -- & -- & --
& 100
& \phantom{--} & -- & -- & \phantom{--}
\\

Qwen 3.6 27B Only
& 91.8 & \textcolor{red!95!black}{20,400} & 75.7
& -- & -- & -- & --
& --
& \phantom{--} & 100 & -- & \phantom{--}
\\

Qwen 3.6 35B Only
& 92.3 & \textcolor{red!95!black}{21,887} & 80.0
& -- & -- & -- & --
& --
& \phantom{--} & -- & 100 & \phantom{--}
\\

\midrule

\textit{Educated Guessing}
& 90.8 & \textcolor{red!95!black}{23,423} & 73.9
& 72.3 & 3.8 & 5.2 & 4.7
& 4.0 & \phantom{--}
& 6.1 & 3.9 & \phantom{--}
\\

\textit{CASCADIA}~\cite{Jiang2026Cascadia}
& 89.8 & \textcolor{red!95!black}{27,857} & 82.1
& 23.1 & 14.8 & 11.6 & 9.5
& 7.2 & \phantom{--}
& 7.6 & 26.2 & \phantom{--}
\\

\textit{MasRouter}~\cite{Yue2025Masrouter}
& \underline{89.8} & \textcolor{green!60!black}{\textbf{16,368}} & \textbf{55.3}
& 88.3 & 1.5 & 2.3 & 0.8
& 1.8 & \phantom{--}
& 3.2 & 2.1 & \phantom{--}
\\

\rowcolor{gray!25}
\textbf{\textsc{ProgRouter}} (\textit{ours})
& \textbf{92.1} & \textcolor{green!60!black}{18,373} & 61.6
& 91.2 & 0.7 & 2.2 & 1.6
& 2.0 & \phantom{--}
& 1.2 & 1.1 & \phantom{--}
\\

\bottomrule
\end{tabularx}

\vspace{-1.5em}
\end{table*}

\vspace{-0.5em}
\section{Experiments}
\vspace{-0.5em}
\subsection{Experimental Setup}

\begin{figure*}[!t]
  \centering
  \includegraphics[width=0.975\linewidth]{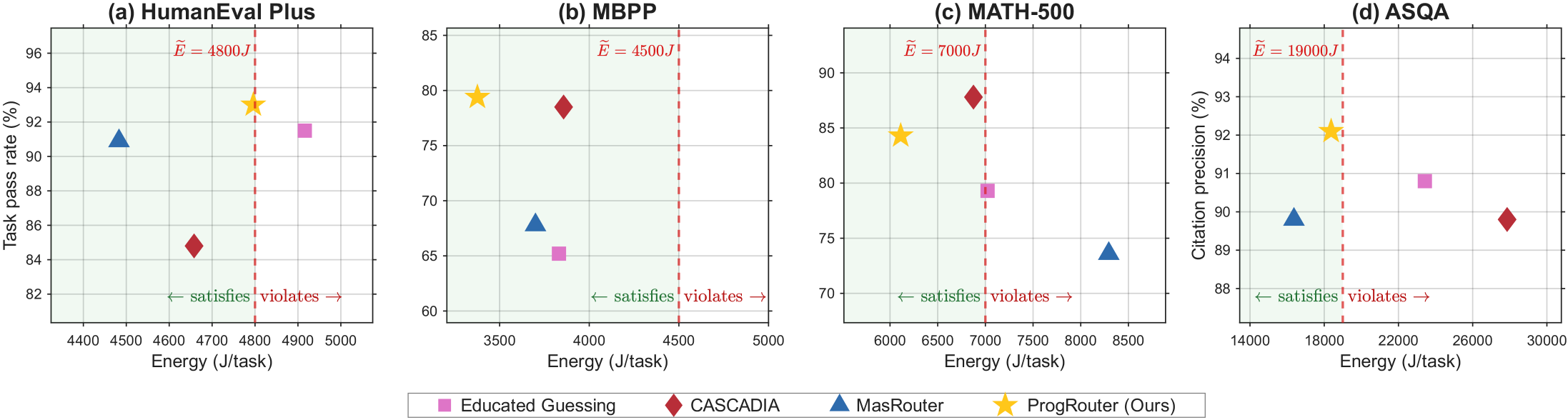}
  \vspace{-0.6em}
  \caption{\textsc{ProgRouter}: Performance-cost tradeoff analysis.}
  \label{fig:pareto}
  \vspace{-1.35em}
\end{figure*}

\begin{figure*}[thbp]
  \centering

  \begin{minipage}{0.38\linewidth}
    \centering
    \vspace{0.85em}
    \includegraphics[width=\linewidth]{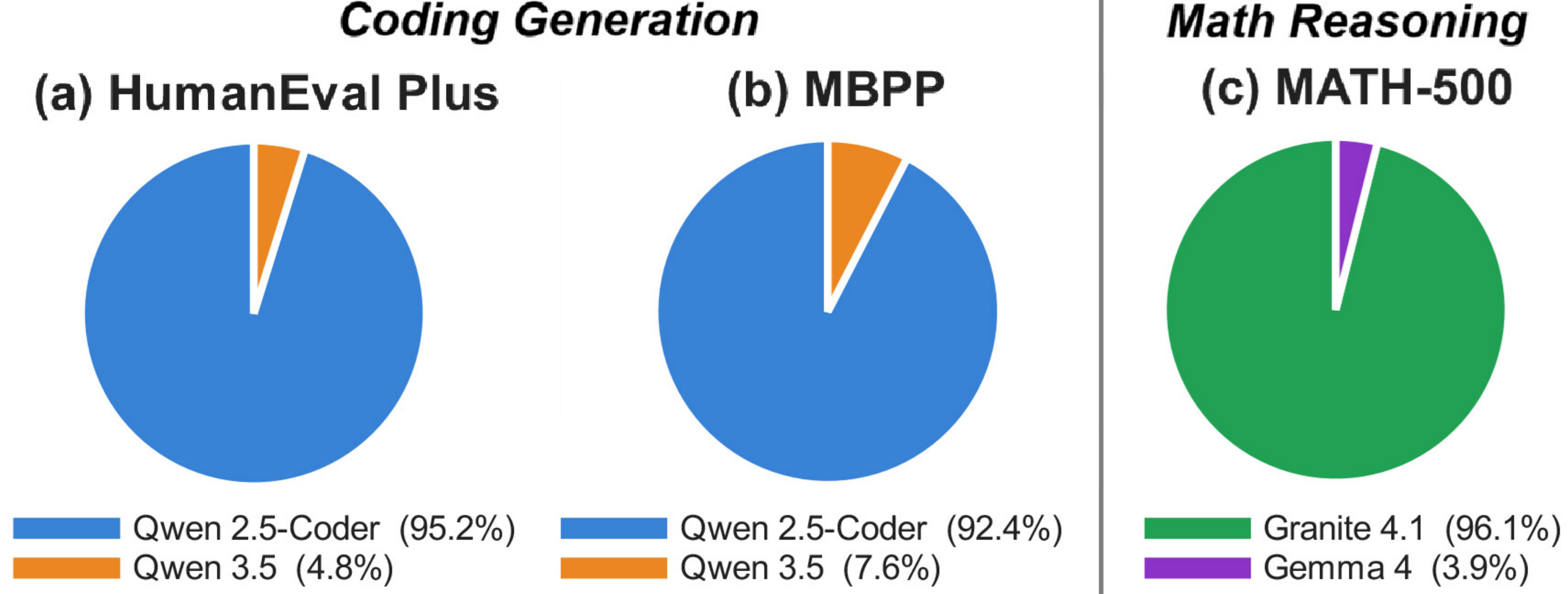}
    \vspace{-1.35em}
    \captionof{figure}{LLM routing distribution by model families.}
    \label{fig:model-family}
  \end{minipage}
  \hspace{0.05\linewidth}
  \begin{minipage}{0.54\linewidth}
    \centering
    \vspace{0.05em}
    \includegraphics[width=\linewidth]{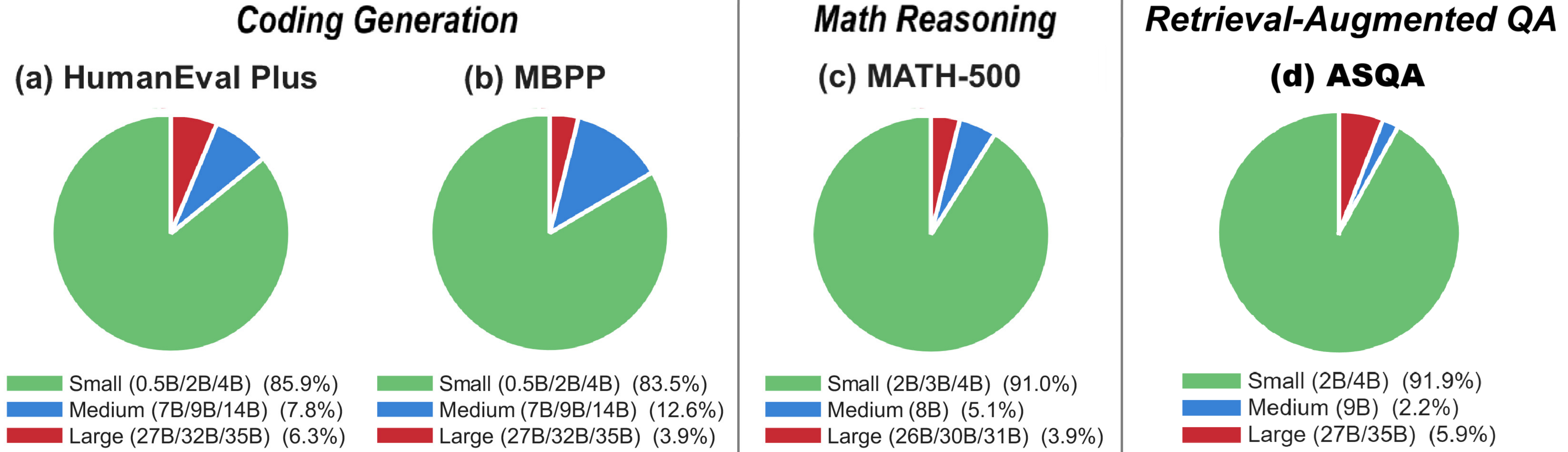}
    \vspace{-1.5em}
    \captionof{figure}{LLM routing distribution by model sizes.}
    \label{fig:model-size}
  \end{minipage}
  \vspace{-1.35em}
\end{figure*}

\textbf{Dataset and Benchmarks.}\!\!
We evaluate \textsc{Prog-} \textsc{Router} in agentic LLM workflows using four widely-adopted benchmarks covering code generation, mathematical reasoning, and retrieval-augmented long-form question answering (QA): HumanEval Plus~\cite{Liu2023Code}, MBPP~\cite{Austin2021Program}, MATH-500~\cite{Hendrycks2021Measuring}, and ASQA~\cite{Stelmakh2022ASQA}. Notably, ASQA evaluates open-ended retrieval-augmented long-form QA, where different LLM agent roles iteratively retrieve evidence, synthesize information, and produce citation-supported answers without unit-test-style feedback. We use the complete set of 164 tasks from HumanEval Plus, along with randomly sampled subsets of 200 coding tasks from MBPP, 200 mathematical reasoning problems from MATH-500, and 100 retrieval-augmented long-form QA tasks from ASQA. For HumanEval Plus, MBPP, and MATH-500, we use task pass rate as the task completion metric, while ASQA is evaluated using citation precision, as it more appropriately measures whether generated responses are supported by relevant and correctly attributed evidence in open-ended retrieval-augmented QA.

\noindent\textbf{Baselines.} We compare \textsc{ProgRouter} with three key baselines: \textbf{(1)}~\textit{MasRouter}~\cite{Yue2025Masrouter}, a query-level one-shot LLM routing approach that assigns each user task to a fixed LLM based on an initial assessment of task complexity before workflow execution; \textbf{(2)}~\textit{CASCADIA}~\cite{Jiang2026Cascadia}, a reactive LLM routing escalation strategy that starts from a small LLM agent and escalates to larger ones when the workflow's task progress stalls; and \textbf{(3)} \textit{Educated Guessing}, which accumulates LLM routing experience online and selects historically best-performing LLMs regardless of actual task progress requirements.

\noindent\textbf{Model Zoo.} For the coding benchmarks (HumanEval Plus and MBPP), the model zoo comprises nine LLMs drawn from two families \textit{Qwen 2.5-Coder} (0.5B/7B/14B/32B) \cite{Hui2024qwen25coder} and \textit{Qwen 3.5} (2B/4B/9B/27B/35B) \cite{Qwen35release2026}. For the math reasoning benchmark (MATH-500), we employ seven LLMs from two families \textit{Granite 4.1} (3B/8B/30B) \cite{Ibmgranite41_2026} and \textit{Gemma 4} (2B/4B/26B/31B) \cite{huggingface_gemma4_2026}. For retrieval-augmented long-form QA benchmark (ASQA), the model zoo consists of seven LLMs from \textit{Qwen 3.5} (2B/4B/9B/27B/35B) \cite{Qwen35release2026} and \textit{Qwen 3.6} (27B~\cite{Qwen36_27B_release2026}/35B~\cite{Qwen36_35B_release2026}).

\subsection{Results Analysis}
Table~\ref{tab:results_all} reports the main benchmark results on HumanEval Plus, MBPP, MATH-500, and ASQA, including task pass/precision rate, energy consumption, workflow execution time, and the LLM routing distribution for each method.

\textbf{Single-model policies provide an unfavorable quality-cost tradeoff.} As shown in Table~\ref{tab:results_all}, most fixed single-model baselines violate the long-term energy-efficiency requirement $\widetilde{E}$, while the few budget-feasible configurations sacrifice task-solving quality. Small LLMs such as Qwen2.5-Coder~0.5B and Granite~4.1~3B incur high cumulative energy because their weak task-solving capacity triggers repeated agent dispatch steps to recover from intermediate failures. Conversely, large LLMs such as Qwen~3.5~35B and Gemma~4~31B exceed $\widetilde{E}$ due to their per-call energy cost, with Gemma~4~31B consuming up to 26,276\,J on MATH-500. This trend also holds in the open-ended ASQA setting, where fixed single-model policies based on Qwen~3.5 and Qwen~3.6 exceed the 19,000\,J energy budget except for the smallest model configuration, which satisfies the budget constraint but achieves lower citation precision. This further demonstrates that neither always-small nor always-large policies are viable under realistic operating constraints, confirming the motivation of \textsc{ProgRouter} that adaptive step-wise LLM agent routing is necessary to balance task quality and operating cost efficiency.

\textbf{\textsc{ProgRouter} achieves the best quality-cost tradeoff among $\widetilde{E}$-satisfying methods.} On HumanEval Plus, \textsc{ProgRouter} attains the highest task pass rate of 93.0\% while remaining within the 4,800\,J budget, outperforming MasRouter ($+2.1$\%) and CASCADIA ($+8.2$\%). On MBPP, \textsc{ProgRouter} achieves the best task pass rate (79.4\%), the lowest energy (3,376\,J), and the shortest execution time (10.3\,s), dominating all baselines on every metric. On MATH-500, \textsc{ProgRouter} delivers the lowest energy (6,112\,J) and the shortest execution time (19.0\,s) among $\widetilde{E}$-satisfying methods, while reaching 84.3\% task pass rate, 3.5 percentage points below
CASCADIA, but at substantially lower operating cost. Beyond code-generation and math reasoning benchmarks, \textsc{ProgRouter} also generalizes to the more open-ended retrieval-augmented QA setting of ASQA, where task progress cannot be specified through clear unit-test-style feedback. On ASQA, \textsc{ProgRouter} achieves the highest citation precision (92.1\%) among all evaluated methods while remaining within the 19,000\,J energy budget, improving over MasRouter (89.8\%) and CASCADIA (89.8\%) by 2.3\%. Figure~\ref{fig:pareto} visualizes this quality--cost tradeoff, where \textsc{ProgRouter} consistently lies on or near the Pareto frontier across all benchmarks. These results validate that the progress-aware LLM routing score Eq.~(\ref{eq:score}), combined with virtual-queue-based budget tracking, enables \textsc{ProgRouter} to navigate the Pareto frontier more effectively than baselines.

\textbf{Adaptive LLM routing baselines exhibit consistent failure modes that \textsc{ProgRouter} avoids.} \textit{Educated Guessing} converges to historically strong but task-agnostic choices (e.g., 98.8\% Granite~4.1~8B on MATH-500), which violates $\widetilde{E}$ on MATH-500 (7,023\,J) and underperforms on harder problems where progress stalls go undetected. \textit{CASCADIA}'s reactive routing escalation incurs unnecessary small-model calls before recovery, leading to a lower task pass rate on HumanEval Plus (84.8\%). \textit{MasRouter}'s routing decisions distribute LLM agent calls without adapting to evolving workflow states and operating cost budgets, resulting in energy violations on MATH-500 (8,294\,J) and a weak MBPP pass rate (67.8\%). In the open-ended ASQA setting, MasRouter also relies heavily on a single dominant small model (88.3\% of calls to Qwen~3.5~2B), which achieves lower citation precision (89.8\%) than \textsc{ProgRouter} (92.1\%) despite lower energy consumption. \textsc{ProgRouter} avoids all these failure modes by continuously estimating real-time task progress and adjusting LLM agent routing decisions to evolving workflow conditions.

\textbf{LLM agent routing distributions reveal adaptive specialization according to task progress requirements.}
As shown in Figures~\ref{fig:model-family}~and~\ref{fig:model-size}, \textsc{ProgRouter} adaptively allocates LLM agent calls according to workflow progress requirements. For code-generation and math reasoning tasks, \textsc{ProgRouter} concentrates the majority of agent dispatch steps on efficient small LLM agents within the dominant model family (e.g., 84.3\% on Qwen2.5-Coder~0.5B for HumanEval Plus and 91.0\% on Granite~4.1~3B for MATH-500), while selectively invoking larger LLM agents when the progress predictor anticipates substantial gains from additional LLM capability. In the open-ended ASQA setting, \textsc{ProgRouter} exhibits a consistent specialization pattern. This demonstrates that \textsc{ProgRouter} does not rely on a fixed model-selection strategy, but instead adapts routing decisions to the characteristics and evolving difficulty of each workflow. This adaptive pattern directly reflects the progress-gap-aware term $V \cdot (1-g(s_t)) \cdot \hat{y}_t(m_t)$ in the LLM routing score (Eq.~\ref{eq:score}), which upweights stronger LLMs only when the workflow has substantial remaining progress to gain.

\vspace{-0.35em}
\section{Conclusion}
\vspace{-0.5em}
This paper presents \textsc{ProgRouter}, an online progress-guided orchestration framework for multi-agent LLM workflows under quality-cost tradeoffs. By combining multi-view task progress scoring with dual-path progress prediction, \textsc{ProgRouter} estimates the marginal progress gain of each candidate LLM agent and leverages it to guide cost-aware, deadline-aware LLM agent routing decisions. Experiments on HumanEval Plus, MBPP, MATH-500 and ASQA benchmarks show that \textsc{ProgRouter} consistently satisfies the long-term operating cost budget while achieving strong task-solving performance and competitive execution time. These results highlight the value of progress-aware online adaptive LLM agent orchestration for multi-agent workflows.

\section*{Limitations}
While \textsc{ProgRouter} demonstrates strong empirical performance, several limitations remain and point to directions for future work. \textit{First}, our evaluation is conducted on four benchmarks covering code generation (HumanEval Plus, MBPP), mathematical reasoning (MATH-500), and retrieval-augmented long-form QA (ASQA). Although these tasks are widely adopted and span multiple agentic domains, the generalization of \textsc{ProgRouter} to other agentic settings, such as open-ended web navigation and tool-augmented QA, remains to be empirically verified. \textit{Second}, the multi-view task progress scorer requires lightweight domain adaptation, where observable workflow milestones and coarse outcome regimes are specified according to task characteristics. Although the current design provides reliable progress signals across diverse benchmarks, automatically learning these progress representations in a fully end-to-end manner remains an interesting direction.

\section*{Ethical Considerations}
\textsc{ProgRouter} is designed to reduce the energy consumption and operating costs of multi-agent LLM workflows, thereby contributing to more sustainable AI serving. Since our method only routes among existing LLMs and does not modify their underlying capabilities, the quality, reliability, and potential biases of generated outputs still depend on the selected models. Standard practices such as careful model selection, output verification, citation checking for retrieval-augmented tasks, and human oversight remain necessary. Because LLM routing decisions are adapted according to task progress, practitioners should monitor routing patterns across different task types to ensure stable service quality and avoid unintended over-reliance on particular models. Our experiments use public benchmarks and do not involve personal data. However, real-world deployments may store user information in workflow states or coordinator ledgers, so appropriate data protection measures, such as access control and data retention policies, should be applied. By improving the operating cost efficiency of agentic LLM workflows, \textsc{ProgRouter} could help make multi-agent LLM systems more sustainable and accessible.


\bibliography{manuscript}

\newpage
\appendix
\section*{Appendices}
Within this supplementary material, we elaborate on the following aspects:
\begin{itemize}[leftmargin=*]
\item Appendix \ref{app:key_notations}: Key notations used in this paper.
\item Appendix \ref{app:dataset_details}: Details of the evaluation datasets and benchmark settings.
\item Appendix \ref{app:baseline_details}: Descriptions of the state-of-the-art baseline methods.
\item Appendix \ref{app:experimental_setup}: Supplementary details of the experimental configuration, energy measurement methodology and evaluation protocol.
\item Appendix \ref{app:ablation}: Ablation analysis results and component-wise evaluations.
\end{itemize}

\section{Key Notations}
\label{app:key_notations}

Table~\ref{tab:notations} summarizes the key notations used throughout this paper.

\begin{table}[!h]
\centering
\small
\setlength{\tabcolsep}{4pt}
\renewcommand{\arraystretch}{1.15}
\caption{Key notations.}
\label{tab:notations}
\vspace{-0.5em}
\begin{tabularx}{\linewidth}{@{}l X@{}}
\toprule
\textbf{Notation} & \textbf{Description} \\
\midrule
\multicolumn{2}{@{}l}{\textit{Multi-Agent LLM Workflow}} \\
\midrule
$w$ & A multi-agent LLM workflow serving user tasks \\
$W$ & The number of user tasks\\
$C$ & Coordinator LLM agent\\
$\mathcal{R}$ & Set of predefined worker LLM agent roles \\
$r_t$ & The worker agent role dispatched at step $t$ \\
$\mathcal{M}$ & Set of available candidate LLMs (model zoo) \\
$m_t$ & The candidate LLM selected at step $t$ \\
$m_t^\star$ & Optimal LLM selected at step $t$ \\
$T_w$ & The number of agent steps for workflow $w$ \\
$t$ & Index of the current agent dispatch step \\
$s_t$ & Workflow state (coordinator ledger) at step $t$ \\
$\mathcal{P}(w)$ & Final task outcome of workflow $w$, $\in[0,1]$ \\
\midrule
\multicolumn{2}{@{}l}{\textit{Operating Costs and Time Constraints}} \\
\midrule
$E(m_t)$ & Operating (energy) cost of invoking LLM $m_t$ \\
$\Gamma(m_t)$ & Execution time incurred by invoking LLM $m_t$ \\
$E_w$ & Per-task operating cost budget for workflow $w$ \\
$\Gamma_w$ & Per-task time budget for workflow $w$ \\
$\widetilde{E}$ & Long-term average operating cost constraint \\
$Q_w$ & Virtual queue tracking cost-constraint violation \\
\midrule
\multicolumn{2}{@{}l}{\textit{Multi-View Task Progress Scoring}} \\
\midrule
$g(s_t)$ & Multi-view task progress score, $\in[0,1]$ \\
$c_t$ & Workflow outcome regime (invalid / recoverable / partial / complete) \\
$b(c_t)$ & Base progress score for regime $c_t$ \\
$r_t$ & Subtask-completion sub-score \\
$d_t$ & Progress-trend sub-score \\
$e_t$ & State-quality sub-score \\
\midrule
\multicolumn{2}{@{}l}{\textit{Dual-Path Task Progress Predictor}} \\
\midrule
$P_{\Theta}$ & Online task progress predictor\\
$\hat{y}_t$ & Predicted progress gain at step $t$ \\
$y_t$ & Realized progress gain, $y_t=g(s_{t+1})-g(s_t)$ \\
$x_t^{\mathrm{str}}$ & Structured (tabular) feature vector at step $t$ \\
$x_t^{\mathrm{sem}}$ & Semantic embedding feature vector at step $t$ \\
$\phi_{\mathrm{str}}, \phi_{\mathrm{sem}}$ & Structured and semantic feature extractors \\
$P_{\mathrm{str}}, P_{\mathrm{sem}}$ & Structured-path and semantic-path regressors \\
$\hat{y}_t^{\mathrm{str}}, \hat{y}_t^{\mathrm{sem}}$ & Progress-gain estimates from the two paths \\
$P_{\mathrm{meta}}$ & Meta-gated learner combining dual-path estimates \\
\midrule
\multicolumn{2}{@{}l}{\textit{Online Routing Decision}} \\
\midrule
$\pi(s_t, r_t)$ & Online LLM routing policy \\
$c_t^{\Gamma}$ & Budget-aware time penalty coefficient \\
$c_t^{E}$ & Budget-aware operating-cost penalty coefficient \\
$\mathrm{score}(m_t)$ & Online LLM routing score at step $t$ \\
\bottomrule
\end{tabularx}
\end{table}

\section{Dataset Details}
\label{app:dataset_details}

We evaluate \textsc{ProgRouter} on four representative benchmarks spanning diverse agentic LLM workflows, including code generation, mathematical reasoning, and retrieval-augmented long-form question answering.

\textbf{HumanEval Plus.}
HumanEval Plus~\cite{Liu2023Code} is an enhanced version of the original HumanEval benchmark for evaluating LLM code generation capability. Each task consists of a programming problem description and requires the LLM agent to generate a function implementation satisfying the provided prompt specification. Compared with the original HumanEval, HumanEval Plus introduces additional hidden test cases to provide a more comprehensive evaluation of functional correctness. In our experiments, we use the complete set of 164 tasks and evaluate performance using task pass rate, where a generated solution is considered successful if it passes all associated test cases.

\textbf{MBPP.}
The Mostly Basic Python Problems (MBPP) benchmark~\cite{Austin2021Program} evaluates LLM agents on basic Python programming tasks. Each task contains a natural-language problem description, a reference implementation, and corresponding test cases for correctness evaluation. MBPP provides a broader set of programming scenarios than HumanEval Plus and evaluates whether generated programs satisfy functional requirements. We randomly sample 200 tasks from the benchmark and measure task completion using pass rate based on automated test execution.

\textbf{MATH-500.}
MATH-500~\cite{Hendrycks2021Measuring} is a mathematical reasoning benchmark containing challenging competition-style problems across multiple mathematical domains. Each problem requires multi-step reasoning and a final answer derivation, making it suitable for evaluating the reasoning capability of agentic workflows. We randomly sample 200 problems and evaluate solutions using task pass rate based on exact answer matching after solution generation.

\textbf{ASQA.}
ASQA (Ambiguous Search Question Answering)~\cite{Stelmakh2022ASQA} is an open-domain retrieval-augmented long-form QA benchmark designed to evaluate whether agentic LLM systems can synthesize accurate and evidence-supported responses for ambiguous questions. Unlike code-generation and mathematical reasoning benchmarks with explicit correctness checks, ASQA requires iterative evidence retrieval, information synthesis, and citation-supported answer generation. Therefore, it provides a complementary evaluation setting for open-ended agentic workflows where task progress must be inferred from intermediate retrieval and reasoning states. We randomly sample 100 ASQA questions and evaluate generated responses using citation precision, which measures whether cited evidence appropriately supports the generated answer.

\section{State-of-the-art Baseline Description}
\label{app:baseline_details}

We compare \textsc{ProgRouter} with three representative state-of-the-art LLM agent routing baselines, covering multi-agent system configuration, cascade-based LLM serving, and online experience-based routing strategies. We additionally evaluate fixed single-model policies as reference baselines to analyze the necessity of adaptive LLM routing under task-solving quality and operating-cost constraints.

\textbf{MasRouter.}
MasRouter~\cite{Yue2025Masrouter} is a learning-based routing framework designed for multi-agent LLM systems.  Unlike conventional single-query LLM routing methods, MasRouter jointly considers collaboration mode selection, agent role allocation, and LLM selection through a cascaded controller architecture. It learns routing decisions based on task-level information and constructs multi-agent configurations that balance task performance and operating cost. In our experiments, we use MasRouter as a representative multi-agent routing baseline that performs LLM selection before or at the beginning of workflow execution based on the available task information.

\textbf{CASCADIA.}
CASCADIA~\cite{Jiang2026Cascadia} is an efficient cascade LLM serving framework that jointly optimizes model deployment and request routing. It formulates cascade serving as a constrained optimization problem, where the deployment component determines resource allocation and parallelism strategies for different LLM candidates, while the routing component optimizes model selection decisions under quality and latency requirements. CASCADIA adopts a bi-level optimization framework consisting of a MILP-based deployment solver and a Chebyshev-guided routing solver to co-optimize system configuration and routing strategies. In our evaluation, CASCADIA routes requests to smaller models for simpler cases and escalates to stronger models when additional capability is required. Unlike \textsc{ProgRouter}, which performs online step-wise routing based on evolving workflow progress, CASCADIA primarily follows a reactive cascade strategy that triggers stronger models when existing execution signals indicate insufficient progress.

\textbf{Educated Guessing.}
Educated Guessing is an online experience-based LLM agent routing baseline that leverages historical execution outcomes to guide future LLM selection. It maintains routing statistics from previous tasks and favors models that have demonstrated strong empirical performance. Unlike \textsc{ProgRouter}, which explicitly estimates workflow progress and marginal progress gain, Educated Guessing does not model the evolving workflow state or incorporate predicted progress improvement into agentic LLM routing decisions.

\textbf{Single-model Baselines.}
We additionally evaluate fixed single-model policies, where the same LLM is selected for all agent dispatch steps throughout workflow execution. These baselines include both lightweight and large-capability models from each benchmark-specific model zoo. They provide reference points for evaluating the effectiveness of adaptive step-wise LLM routing and quantify the tradeoff between model capability, task-solving performance, and operating cost.

\section{Supplementary Experimental Setup}
\label{app:experimental_setup}

\subsection{System Configuration} 

We implement \textsc{ProgRouter} on top of AG2 0.11.3~\cite{Wu2023autogen}, an open-source multi-agent LLM framework, and serve the routed LLM agents using Ollama 0.24.0 on an NVIDIA RTX PRO 6000 Blackwell workstation. In all evaluated benchmarks, the coordinator LLM agent is powered by Qwen3-Coder-Next 80B-A3B \cite{Cao2026Qwen3CoderNext}. The long-term averaging energy budget $\widetilde{E}$ is calibrated separately for each benchmark to reflect task complexity, and is set to $\widetilde{E}=4,800$~J for HumanEval Plus, $\widetilde{E}=4,500$~J for MBPP, $\widetilde{E}=7,000$~J for MATH-500, and $\widetilde{E}=19,000$~J for ASQA.

\subsection{Energy Measurement Methodology}

The energy consumption of LLM agent serving is measured using NVIDIA NVML with a 100~ms sampling interval. Unlike token-count or FLOP-based proxies, our energy measurement is based on physical GPU power measurements. During calibration, we integrate the active GPU power over each execution segment:
\begin{equation}
E=\int_{t_{\mathrm{start}}}^{t_{\mathrm{end}}}(P(t)-P_{\text{idle}})\,dt,
\end{equation}
where $P(t)$ is the GPU board power reported by NVML and $P_{\mathrm{idle}}$ is the measured idle GPU power. The resulting value represents the incremental GPU energy attributable to the workload.

We separately profile LLM model loading, prompt prefill, token generation, worker/coordinator LLM agent execution, and tool execution. For each LLM size in the zoo, we calibrate a per-call energy estimator:
\begin{equation}
\begin{aligned}
E_{\mathrm{estimated}}
&= I_{\mathrm{load}}E_{\mathrm{load}} + E_{\mathrm{prefill}} + e_{\mathrm{decode}}N_{\mathrm{gen}} .
\end{aligned}
\end{equation}
where $I_{\mathrm{load}}$ indicates whether the call incurs a cold model load, $E_{\mathrm{prefill}}$ captures prompt-processing energy based on the actual context length, and $N_{\mathrm{gen}}$ denotes the generated token count. This formulation characterizes prompt processing and autoregressive token generation.

The reported GPU energy includes both worker-agent and coordinator-agent LLM inference, covering model loading, prompt prefill, context-dependent computation, token decoding, KV-cache activity, and GPU runtime overhead. CPU-side orchestration, host memory, storage, and network communication energy are considered negligible and outside the energy measurement boundary.

\subsection{Online Evaluation Protocol}
\label{app:online_protocol}

\textsc{ProgRouter} performs online learning during workflow execution, where the task progress predictor is trained and optimized on the fly using realized task progress feedback collected from previous routing decisions. To evaluate this online adaptation process, we construct each benchmark evaluation as a sequential task stream. Each benchmark task set is randomly shuffled using a fixed random seed, and each task is processed once in the resulting order without replay. The same task stream construction is used for all compared methods.

At the beginning of evaluation, the task progress predictor starts from a cold initialization without pretrained parameters or offline training samples. During the warm-up phase, \textsc{ProgRouter} uses $\epsilon$-greedy exploration to collect LLM agent routing trajectories and corresponding realized task progress signals as training targets. After a worker LLM executes a selected workflow step, the resulting workflow state is evaluated by the task progress scorer, producing the realized progress gain:
\begin{equation}
y_t = g(s_{t+1}) - g(s_t),
\end{equation}
which is used as the training target for optimizing the task progress predictor. This exploration procedure is performed online, and the resulting progress feedback is available only after the selected workflow step has been executed.

The task progress predictor's training set buffer, learned parameters, and virtual queue state persist throughout the task stream. The task progress predictor is periodically updated as additional routing trajectories become available. We define stabilization adaptively based on predictor maturity. Specifically, stabilization is reached when the training buffer contains sufficient realized-progress samples and the exploration probability decreases below the predefined threshold. After stabilization, the learned routing policy is evaluated on the remaining task stream. The reported main results correspond to the steady-state evaluation after stabilization and therefore measure the performance of the adapted LLM agent routing policy.

\section{Ablation Analysis Results}
\label{app:ablation}

We conduct additional ablation studies on the HumanEval Plus benchmark to analyze the contribution of individual components in \textsc{ProgRouter}. The ablations examine two aspects separately: (1) the internal design of the dual-path progress predictor, including the structured path, semantic path, and adaptive meta-learner; and (2) the online routing components, including the multi-view progress scorer, budget-aware routing objective, and virtual queue mechanism. All online routing ablations adopt the same model zoo, workflow configuration, energy budget, and evaluation protocol (Appendix~\ref{app:online_protocol}).

\subsection{Ablation of Progress Prediction Components}

The dual-path task progress predictor is designed to capture complementary workflow information from structured execution features and semantic workflow states. We first evaluate each prediction component through offline prediction accuracy analysis. Table~\ref{tab:ablation_predictor} reports the prediction error of different predictor variants.

\begin{table}[t]
\centering
\small
\caption{Offline ablation of progress predictor components on HumanEval Plus. Lower MAE indicates more accurate progress prediction.}
\label{tab:ablation_predictor}

\begin{tabular}{lcc}
\toprule
\textbf{Predictor Variant} & \textbf{MAE} & \textbf{$\Delta$ MAE} \\
\midrule
Structured path only & 0.0967 & +0.0247 \\
Semantic path only & 0.0788 & +0.0068 \\
Dual paths, mean-combined & 0.0843 & +0.0123 \\
\textbf{Dual paths + meta-learner} & \textbf{0.0720} & \textbf{0} \\
\bottomrule
\end{tabular}
\end{table}

The results show that both prediction paths provide complementary information. Removing either path increases prediction error, indicating that structured workflow signals and semantic state representations capture different aspects of task evolution. The semantic path achieves lower error than the structured path alone, suggesting that intermediate workflow descriptions contain valuable information beyond explicit execution statistics. However, simply averaging the two prediction paths is inferior to the adaptive meta-learner, demonstrating that the contribution of each path varies across workflow states. The meta-learner dynamically combines structured and semantic predictions to achieve the best task progress estimation accuracy.

\subsection{Ablation of Online Routing Components}

We further evaluate how individual routing components contribute to end-to-end workflow performance. Table~\ref{tab:ablation_routing} reports online evaluation results on HumanEval Plus under the same long-term energy constraint.

\begin{table}[t]
\centering
\small
\caption{Ablation of online routing components on HumanEval Plus.}
\label{tab:ablation_routing}

\begin{tabular}{lcc}
\toprule
\textbf{Configuration} & 
\textbf{Pass (\%)} & 
\textbf{Energy (J)} \\
\midrule
\textbf{Full ProgRouter} & \textbf{93.0} & \textbf{4,796} \\
w/o predictor & 89.0 & 4,785 \\
Structured path only & 90.9 & 4,400 \\
Semantic path only & 87.2 & 4,784 \\
w/o multi-view scorer & 90.2 & 4,486 \\
w/o budget penalty & 87.8 & 4,252 \\
w/o queue penalty & 92.7 & 4,406 \\
Naive task progress/cost & 17.7 & 7,797 \\
\bottomrule
\end{tabular}
\end{table}

Removing the progress predictor reduces the pass rate from 93.0\% to 89.0\%, demonstrating that estimating future progress gain is important for selecting appropriate LLM agents. Using only one prediction path also decreases performance, consistent with the offline prediction results. In particular, the semantic-only variant suffers a larger degradation because it lacks explicit structured workflow signals.

The multi-view task progress scorer also contributes to LLM agent routing effectiveness. Replacing it with only the test-pass-rate view reduces the pass rate to 90.2\%, showing that intermediate signals such as subtask completion, progress trends, and workflow-state quality provide useful information beyond final outcome estimation.

The budget-aware penalty plays an important role in balancing task progress and operating cost. Removing this component decreases performance to 87.8\%, despite reducing energy consumption, because the LLM agent router can no longer effectively account for heterogeneous models' operating costs when selecting candidate LLM agents. The virtual queue mechanism has a smaller effect on this finite evaluation stream (92.7\% versus 93.0\%), which is consistent with its intended purpose of controlling long-term budget deviation rather than improving individual task performance.

\subsection{Progress Signal Alone is Insufficient}

To distinguish the contribution of the complete routing objective from the task progress predictor itself, we additionally evaluate a naive task progress-per-cost routing strategy. This baseline uses the same predicted progress gain as \textsc{ProgRouter} but greedily selects the model with the largest predicted progress improvement per unit normalized cost, without considering remaining progress gap, accumulated budget deviation, or long-term routing consequences.

The naive strategy achieves only 17.7\% pass rate while consuming 7,797\,J, substantially worse than the full \textsc{ProgRouter} system (93.0\% pass rate and 4,796\,J). The failure occurs because immediate progress-per-cost optimization favors inexpensive but insufficiently capable LLMs, causing workflow stalls and repeated recovery attempts. These results demonstrate that accurate progress prediction alone is insufficient; effective LLM agent routing requires jointly considering predicted progress gain, remaining task difficulty, and long-term budget constraints.

Overall, the above ablation results confirm that the advancements of \textsc{ProgRouter} arise from the interaction of multiple algorithmic components. The dual-path task progress predictor provides accurate progress estimation; the multi-view task progress scorer captures workflow evolution; and the budget-aware online LLM agent routing objective converts these signals into effective LLM agent selection decisions.

\end{document}